\documentclass{article}

\usepackage{authblk}
\usepackage{orcidlink}
\usepackage{cite}
\usepackage{amsmath,amssymb,amsfonts}
\usepackage{algorithmic}
\usepackage{graphicx}
\usepackage{algorithm,algorithmic}
\usepackage{hyperref}
\hypersetup{hidelinks=true}
\usepackage{textcomp}
\usepackage{glossaries}
\usepackage{booktabs}
\usepackage{array}
\usepackage{tcolorbox}
\usepackage{graphicx}
\usepackage{subcaption} 
\usepackage{enumitem}
\usepackage{listings}
\usepackage{xcolor}
\usepackage{caption}

\definecolor{background}{HTML}{FAFAFA}

\lstdefinelanguage{Swift}
{
  morekeywords={
    open,catch,@escaping,@main,nil,throws,throw,rethrows,rethrow,try,func,if,then,else,for,in,while,do,switch,case,where,break,continue,fallthrough,return,async,await,guard,
    typealias,associatedtype,struct,class,enum,protocol,var,func,let,get,set,willSet,didSet,import,inout,init,deinit,extension,
    subscript,prefix,operator,infix,postfix,precedence,associativity,left,right,none,convenience,dynamic,
    final,lazy,mutating,nonmutating,optional,override,required,static,unowned,safe,weak,internal,
    private,public,is,as,self,unsafe,dynamicType,true,false,nil,Type,Protocol,
    @Parameter
  },
  morecomment=[l]{//}, 
  morecomment=[s]{/*}{*/}, 
  morestring=[b]", 
  breaklines=true,
  escapeinside={\%*}{*)},
  numbers=left,
  captionpos=b,
  breakatwhitespace=true,
  basicstyle=\linespread{1.0}\ttfamily\footnotesize, 
}

\definecolor{keyword}{HTML}{BA2CA3}
\definecolor{string}{HTML}{D12F1B}
\definecolor{identifier}{HTML}{02638C}
\definecolor{number}{HTML}{2729D8}

\definecolor{comment}{rgb}{0.4470588235,0.4980392157,0.5490196078}

\lstdefinestyle{swiftstyle}{
	basicstyle=\ttfamily\small,
    language=Swift,
    numbers=left,
    numberstyle=\tiny,
    stepnumber=1,
    numbersep=8pt,
    showstringspaces=false,
    postbreak=\raisebox{0ex}[0ex][0ex]{\ensuremath{\color{gray}\hookrightarrow\space}},
    breaklines=true,
    frame=leftline,
    inputencoding=utf8,
    extendedchars=\true,
    columns=fullflexible,
    keepspaces=true,
    tabsize=4,
    keywordstyle=\color{keyword},
    stringstyle=\color{string},
    commentstyle=\color{comment},
    showspaces=false,        
    showstringspaces=false,  
    extendedchars=true,      
    backgroundcolor=\color{background}
}

\usepackage[font={small}]{caption}
\colorlet{punct}{red!60!black}
\definecolor{delim}{RGB}{20,105,176}
\colorlet{numb}{magenta!60!black}

\lstdefinelanguage{json}{
    basicstyle=\ttfamily\small,
    numbers=left,
    numberstyle=\tiny,
    stepnumber=1,
    numbersep=8pt,
    showstringspaces=false,
    postbreak=\raisebox{0ex}[0ex][0ex]{\ensuremath{\color{gray}\hookrightarrow\space}},
    breaklines=true,
    frame=leftline,
    morecomment=[s]{/*}{*/}, 
    morestring=[b]", 
    backgroundcolor=\color{background},
    literate=
        *{0}{{{\color{numb}0}}}{1}
        {1}{{{\color{numb}1}}}{1}
        {2}{{{\color{numb}2}}}{1}
        {3}{{{\color{numb}3}}}{1}
        {4}{{{\color{numb}4}}}{1}
        {5}{{{\color{numb}5}}}{1}
        {6}{{{\color{numb}6}}}{1}
        {7}{{{\color{numb}7}}}{1}
        {8}{{{\color{numb}8}}}{1}
        {9}{{{\color{numb}9}}}{1}
        {:}{{{\color{punct}{:}}}}{1}
        {,}{{{\color{punct}{,}}}}{1}
        {\{}{{{\color{delim}{\{}}}}{1}
        {\}}{{{\color{delim}{\}}}}}{1}
        {[}{{{\color{delim}{[}}}}{1}
        {]}{{{\color{delim}{]}}}}{1},
}

\makeglossaries

\newacronym{MVP}{MVP}{minimum viable product}
\newacronym{AI}{AI}{artificial intelligence}
\newacronym{PHI}{PHI}{protected health information}
\newacronym{PHR}{PHR}{personal health record}
\newacronym{LLM}{LLM}{large language model}
\newacronym{EHR}{EHR}{electronic health record}
\newacronym{FHIR}{FHIR}{Fast Healthcare Interoperability Resources}
\newacronym{IoT}{IoT}{Internet of Things}
\newacronym{API}{API}{application programming interface}
\newacronym{DSL}{DSL}{domain-specific language}
\newacronym{JSON}{JSON}{JavaScript object notation}
\newacronym{URI}{URI}{uniform resource identifier}
\newacronym{UML}{UML}{unified modeling language}
\newacronym{mDNS}{mDNS}{Multicast Domain Name Service}
\newacronym{DNS-SD}{DNS-SD}{DNS Service Discovery}
\newacronym{IP}{IP}{Internet Protocol}
\newacronym{SSL}{SSL}{Secure Sockets Layer}
\newacronym{TLS}{TLS}{Transport Layer Security}
\newacronym{ANE}{ANE}{Apple Neural Engine}
\newacronym{GGUF}{GGUF}{GPT-Generated Unified Format}
\newacronym{SPM}{SPM}{Swift Package Manager}
\newacronym{CA}{CA}{Certificate Authority}
\newacronym{SoC}{SoC}{System on a Chip}
\newacronym{RAG}{RAG}{Retrieval-Augmented Generation}
\newacronym{PAEHR}{PAEHR}{Patient-accessible electronic health record}
\newacronym{LEP}{LEP}{limited English proficiency}
\glsdisablehyper

\begin{document}

\title{Dynamic Fog Computing for Enhanced LLM Execution in Medical Applications}

\author[1,2,*]{Philipp Zagar~\orcidlink{0009-0001-5934-2078}~}
\author[1]{Vishnu Ravi~\orcidlink{0000-0003-0359-1275}~}
\author[1]{Lauren Aalami~\orcidlink{0009-0007-7132-5362}~}
\author[2]{Stephan Krusche~\orcidlink{0000-0002-4552-644X}~}
\author[1]{Oliver Aalami~\orcidlink{0009-0001-5934-2078}~}
\author[1]{Paul Schmiedmayer~\orcidlink{0000-0002-8607-9148}~}

\affil[1]{Stanford Mussallem Center for Biodesign, Stanford, 94305, United States}
\affil[2]{Technical University of Munich, Munich, 80333, Germany}
\affil[*]{Corresponding Author: zagar@stanford.edu}

\maketitle

\begin{abstract}
The ability of \glspl{LLM} to transform, interpret, and comprehend vast quantities of heterogeneous data presents a significant opportunity to enhance data-driven care delivery. 
However, the sensitive nature of \gls{PHI} raises valid concerns about data privacy and trust in remote \gls{LLM} platforms.
In addition, the cost associated with cloud-based \gls{AI} services continues to impede widespread adoption.
To address these challenges, we propose a shift in the \gls{LLM} execution environment from opaque, centralized cloud providers to a decentralized and dynamic fog computing architecture. 
By executing open-weight \glspl{LLM} in more trusted environments, such as the user’s edge device or a fog layer within a local network, we aim to mitigate the privacy, trust, and financial challenges associated with cloud-based \glspl{LLM}. 
We further present \textit{SpeziLLM}, an open-source framework designed to facilitate rapid and seamless leveraging of different \gls{LLM} execution layers and lowering barriers to \gls{LLM} integration in digital health applications. 
We demonstrate \textit{SpeziLLM's} broad applicability across six digital health applications, showcasing its versatility in various healthcare settings.
\end{abstract}

\newpage
\section{Introduction}
\label{sec:introduction}

The convergence of digital technology and healthcare has revolutionized medical monitoring and intervention, generating vast amounts of data through \glspl{EHR}, wearable devices, and digital health applications.
Used responsibly, this data can transform health care delivery, increasing patient engagement and improving outcomes~\cite{tapuriapatientehiaccess2021, krusehitoutcomes2018}.
To empower patients and digital health innovators to harness this data, the United States’ 21st Century Cures Act, mandates \gls{EHR} data accessibility via \gls{FHIR} \glspl{API}~\cite{lye2018curesact}.
Nevertheless, challenges in efficiently leveraging this data persist.

\glspl{LLM} have the potential to democratize the vast quantities of accessible health information, advancing healthcare objectives, reducing costs, and improving patient outcomes~\cite{thirunavukarasu2023large, shah2023creation}.
Such models generate human-like text from provided information, effectively bridging the gap between raw data and user interpretation and transforming vast amounts structured or unstructured data into human-legible insights.
\Glspl{LLM} can answer questions, summarize, paraphrase, and interpret text, outperforming human experts in certain contexts~\cite{vanveen2023clinical}.

Efficacy notwithstanding, the adoption of \glspl{LLM} in healthcare contexts raises key data privacy and trust concerns~\cite{yuan2023llm}.
The management of sensitive personal health data in cloud-hosted \gls{LLM} execution environments has profound transparency, regulatory, and security implications ~\cite{yuan2023llm, weidinger2021ethical}.
Scaling \glspl{LLM} is also resource-heavy, with high hardware costs that may deter would-be adopters.

To address these limitations, we present a paradigm shift in the digital health approach to \glspl{LLM}: a fog-computing based architecture that dynamically relocates model execution closer to user devices~\cite{bonomi2012fog} (see \autoref{fig:proposed-fog-llm-architecture}).
Our approach creates the foundation for \gls{LLM} inference environments on widely available, decentralized computing assets, such as users' mobile phones, laptops, or existing computing resources within secure, isolated hospital and clinic networks.
We propose using multiple \gls{LLM} execution environments to account for the limited capabilities of low-power resources running smaller models and the trust and financial concerns associated with remote cloud computing.

\begin{tcolorbox}[title=Research Question]
How can a dynamic fog computing architecture be used to dispatch \gls{LLM} inference tasks across decentralized edge, fog, and cloud environments to address privacy, trust, and financial concerns in digital health applications?
\end{tcolorbox}

To complement our \gls{LLM} fog architecture, we additionally present \textit{SpeziLLM}, an open-source software framework containing all the necessary tools for integrating \gls{LLM}-related functionality
in digital health applications. We demonstrate SpeziLLM’s
broad applicability across six digital health applications, showcasing its versatility in various healthcare settings, and assess its usability, functionality, and adaptability via a survey of student mobile health developers.
The LLM space is fast-developing, so SpeziLLM was designed to be \gls{LLM}-agnostic--able to integrate with various \glspl{LLM} without being tied to any specific one. As such, this paper does not evaluate the output quality of particular \glspl{LLM}.
\section{Background}
\label{sec:background}

\Glspl{PAEHR} have been shown to improve patient outcomes, improving engagement, self-management, and informed decision-making\cite{richwineehidisparities2023, tapuriapatientehiaccess2021}.
However, \gls{LEP} and health literacy disparities may hinder \gls{PAEHR} utilization~\cite{sundellhealthliteracy2022, baileydigitalliteracy2014}.
\glspl{LLM}' sophisticated querying capabilities have the potential to mitigate these barriers, enabling patients to generate context-aware insights into their health data at any degree of complexity and in various languages.
Preliminary validations have established the efficacy of \glspl{LLM} such as OpenAI's GPT in this context~\cite{johnson2023assessing, nori2023capabilities}.
One such demonstration is the open-source \textit{LLMonFHIR} application~\cite{schmiedmayer2023llmonfhir}, designed to facilitate a "dialogue" with \gls{FHIR} health information~\cite{schmiedmayer2024llmonfhir}.

Although \glspl{LLM} frequently provide accurate information, their tendency to generate erroneous content--"\textit{hallucinations}"--prevents them from serving as infallible or singular sources of truth in production environments, especially in high-stakes care delivery contexts.
Ongoing investment in \gls{AI} alignment and hallucination mitigation is reducing their frequency~\cite{ziwei2023hallucination}.
Accordingly, \textit{SpeziLLM} is \gls{LLM}-agnostic, allowing the continuous integration of newer, lower-hallucinogenic models~\cite{xu2024hallucination}.

\textbf{Edge computing}, a distributed paradigm, brings computation and data storage closer to data sources, enhancing trust, security, and privacy by processing data locally~\cite{keyan2020edge}.
Adapting compute-intensive \glspl{LLM} for edge computing can be challenging, due to the disparity in computing resources between consumer-grade mobile edge devices and specialized cloud-based infrastructures.
Edge devices’ limited computational power, resulting from physical dimensions, heat dissipation, battery life, and cost constraints, restricts memory allocation and GPU capabilities.
Ongoing efforts to overcome these limitations and enhance mobile inference performance include the introduction of more compact \glspl{LLM} and model compression techniques (4-bit or 1-bit quantization)~\cite{frantar2023gptq, gunter2024appleintelligence, alizadeh2024llm, mckinzie2024mm1, moniz2024realm}.

\textbf{Fog computing} leverages the strengths of both edge and cloud computing~\cite{bonomi2012fog, iorga2018fog}.
This architectural paradigm brings the substantial computational capabilities of central instances closer to consumers in a distributed fashion, lowering latency and optimizing network resources~\cite{bonomi2012fog}.
Fog computing extends the cloud computing paradigm to the network’s edge, offering dynamic dispatch, agile computing resource allocation, increased efficiency, and enhanced trust in the execution environment~\cite{li2018data}.
Central to this approach are fog nodes: heterogeneous devices stationed near the network’s edge~\cite{vaquero2014finding}.
Fog computing architectures are divided into three layers~\cite{iorga2018fog}:

\begin{itemize}[leftmargin=*, nosep]
    \item \textbf{Edge Layer (Local):} Low-power \Gls{IoT} and end-user devices at the network’s edge where data is generated and utilized.
    \item \textbf{Fog Layer:} Positioned between the cloud and the edge, fog nodes with substantial computational power process data nearer to the source--a more trusted environment than distant servers~\cite{das2023reviewfog}.
    \item \textbf{Cloud Layer (Remote):} Centralized units with massive computational resources, raising privacy, trust, and financial concerns.
\end{itemize}

\noindent Establishing a fog computing system, though advantageous, is technically complex, requiring significant time and expertise to configure and maintain~\cite{das2023reviewfog}.
Additional challenges include interoperability between components, heterogeneity, quality of service, network communication, and resource management~\cite{ahmadi2021fog}.

Fog computing has enhanced real-time health monitoring by serving as an intermediate layer between edge devices and cloud servers~\cite{ahmadi2021fog}.
These applications improve computing efficiency, data security, and bandwidth utilization, enabling low-latency responses and effective management of physiological data, underscoring the potential of fog computing to improve patient care~\cite{dubey2017fog, jain2020adoption}.
\section{Architecture}
\label{sec:architecture}

\begin{figure}[htbp]
    \centering
    \includegraphics[width=\linewidth]{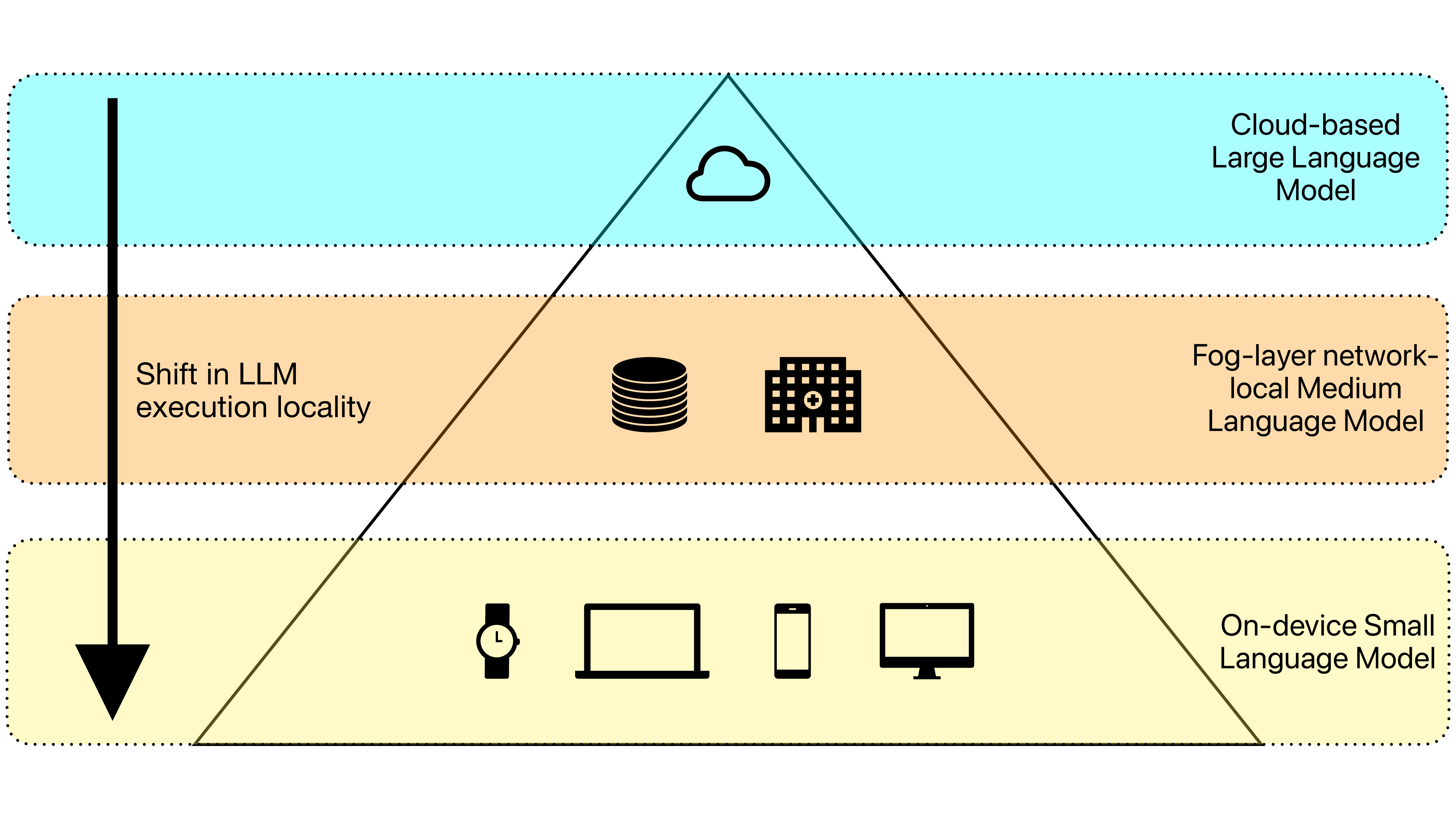}
    \caption{The proposed transition of the \gls{LLM} inference environment from cloud-based platforms to the immediate proximity of the user's device.}
    \label{fig:proposed-fog-llm-architecture}
\end{figure}

Our goal is to transparently and dynamically shift the inference environment of \glspl{LLM} closer to the user's device, depending on cardinality, trust, and computational power for each layer (\autoref{fig:proposed-fog-llm-architecture}).
We aim to provide a uniform and interchangeable interface for interacting with \glspl{LLM}, regardless of execution locality.

\begin{figure}[htbp]
    \centering
    \includegraphics[width=\linewidth]{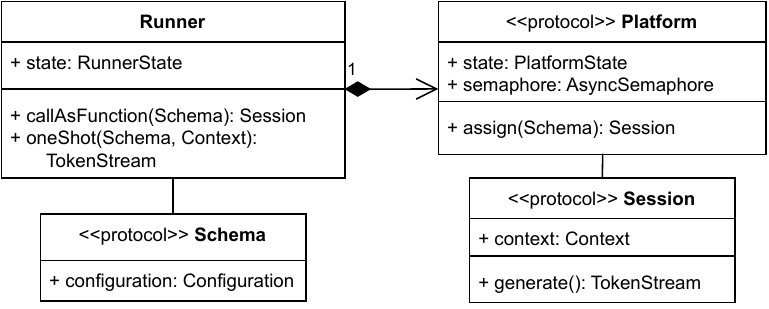}
    \caption{Mental model of all \gls{LLM} interactions as a \gls{UML} class diagram.}
    \label{fig:spezillm_class_diagram}
\end{figure}

In order to establish a shared mental model, we propose a set of terminologies and concepts applicable to \gls{LLM} interactions (\autoref{fig:spezillm_class_diagram}).
This model contains four major components.
A \textbf{Schema} serves as a blueprint for a specific model configuration, representing parameters such as model type or temperature.
The schema, once initialized, is immutable, allowing for key consistency and side effect guarantees.
It does not contain a variable or in-flight inference state.
The schema is passed to a centralized \textbf{Runner} component, which internally delegates and assigns it to a specific local, fog, or remote \textbf{Platform}.
The runner is configured with various platforms, representing a system's different \gls{LLM} inference layers.
A platform is responsible for setting up the \gls{LLM} execution environment and ensuring concurrency-safe access to shared resources.
The platform turns the schema into a concrete \textbf{Session} that acts as the \gls{LLM} in execution, housing the current context and in-flight generation state and performing the actual inference.

\subsection{Retrieval-Augmented Generation Capabilities}

Limited \gls{LLM} context size poses a key challenge for \gls{LLM}-based interactions with large amounts of data.
Injecting all relevant data into the entire context window may also prove financially burdensome, given that the pricing of \gls{LLM} inference is typically based on a per-token cost for \gls{LLM} output and input.

To mitigate these challenges, major \gls{LLM} service providers like OpenAI and Anthropic and openly available \glspl{LLM} like Llama3~\cite{dubey2024llama3} have introduced \textit{function calling} mechanisms~\cite{kim2024llm},
 specific instantiations of \gls{RAG}~\cite{lewis2021retrievalaugmented} that enable \glspl{LLM} to have structured and reliable interaction with external systems.

\begin{figure}[htbp]
    \centering
    \includegraphics[width=0.9\linewidth]{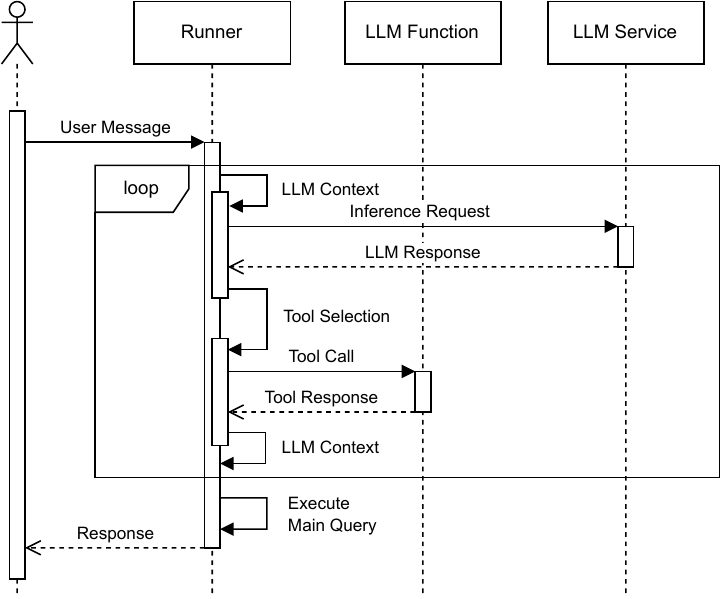}
    \caption[\gls{UML} sequence diagram of the typical \gls{LLM} function calling mechanism.]{\gls{UML} sequence diagram of the typical \gls{LLM} function calling mechanism integrated into our proposed mental model.}
    \label{fig:spezillm-function-calling}
\end{figure}

We integrated these mechanisms as a core feature in our proposed system architecture.
We provide developers with a convenient and declarative \gls{LLM}-agnostic \gls{DSL}~\cite{dsl} that abstracts technical complexities and state management (\autoref{fig:spezillm-function-calling}).
The runner (or, more specifically, the active session) is initially configured with a collection of \textbf{LLM Function}\textbf{s}.
Upon submission of the request message into the runner context, the inference process on the \textbf{LLM Service} is executed.
The service may return a selection of \textit{tools}, including the function name and encoded parameters, in addition to a humanly legible output stream of tokens.
The runner identifies the configured \gls{LLM} functions, injects the supplied parameters, and executes them concurrently.
The results are seamlessly reintegrated into the runner’s context.

\subsection{Dynamic \gls{LLM} Task Dispatching}

The fog layer provides significant decentralized computing resources near the user's edge device, where health data resides and prompts are devised.
Edge devices use dynamic dispatch to allocate inference tasks to the fog node with the highest proximity measurement, thus accurately identifying the most capable and reliable computing resource.

\begin{figure}[htbp]
    \centering
    \includegraphics[width=0.9\linewidth]{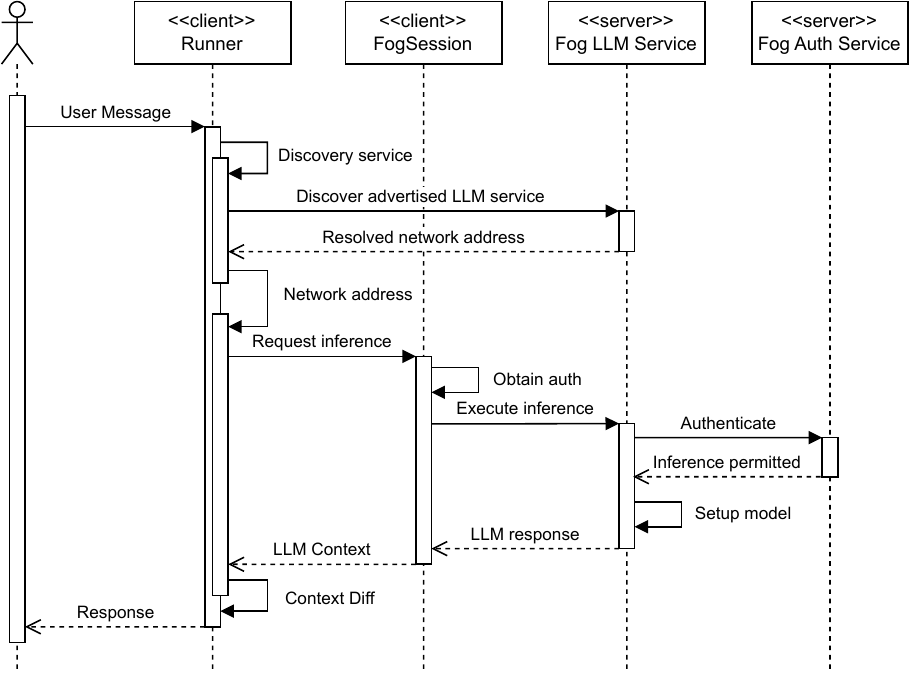}
    \caption[Typical procedure for \gls{LLM} inference job execution within the fog layer.]{Typical procedure for \gls{LLM} inference job execution within the fog layer and our established mental model as a \gls{UML} sequence diagram.}
    \label{fig:llm-fog-inference}
\end{figure}

\autoref{fig:llm-fog-inference} illustrates the process by which an \gls{LLM} inference job is assigned to a computing resource within the fog layer.
The process requires two components: A fog node that advertises \gls{LLM} inference services (acting as the server) and a client that discovers and consumes the \gls{LLM} resource.
Upon receipt of a user message, the runner discovers \textbf{Fog LLM Service}\textbf{s} available within the local network. 
Once a service is discovered, the \gls{LLM} \textbf{FogSession} dispatches an inference job with proper authorization credentials to the previously discovered fog \gls{LLM} service via a secure connection.
If the use of that resource is permitted by the \textbf{Fog Auth Service}, the fog node will set up the language model and stream the content back to the client, who displays the response.

\section{SpeziLLM}
\label{sec:spezillm}

Based on the fog computing \gls{LLM} architecture described in \autoref{sec:architecture}, we propose a platform enabling developer access to \glspl{LLM} across edge, fog, and cloud layers.
Our goal is to empower developers to integrate \glspl{LLM} securely, privately, and cost-efficiently, simplifying the complexities of the decentralized architecture for digital health innovators.
To that end, we introduce \textit{SpeziLLM}\footnote{\url{https://github.com/StanfordSpezi/SpeziLLM/}}~\cite{schmiedmayer2023spezillm}:
an open-source, MIT-licensed Swift software framework offering modularized, ready-to-use \gls{LLM} capabilities that can be combined to suit a breadth of patient care and clinical research goals.
SpeziLLM is embedded in the \textit{Stanford Spezi} ecosystem~\cite{aalami2023cardinalkit, schmiedmayer2023spezi},
which enables the rapid development of component-based, interoperable, and reusable digital health applications.
As a first demonstration, SpeziLLM supports all major Apple operating systems, allowing seamless \gls{LLM} integration across iOS, macOS, and visionOS.
The framework additionally leverages Apple's hardware and software ecosystem, integrating with HealthKit and utilizing Apple Silicon and \textit{Metal}\footnote{\url{https://developer.apple.com/metal/}} acceleration.

The SpeziLLM framework encompasses a suite of convenience components essential for most \gls{LLM} applications, including context data models, onboarding facilitators, \gls{LLM} chat interfaces, and state and error management mechanisms.
All layers are based on the shared mental model presented in \autoref{sec:architecture}, facilitating a reusable and extensible ecosystem across the various execution layers.

\subsection{Cloud or remote layer}
\label{subsec:architecure-remote}

OpenAI and its \gls{LLM} inference \gls{API} have become the blueprint among cloud service providers.
SpeziLLM's cloud layer has been designed for interaction with remote \glspl{LLM} via the OpenAI \gls{API} schema, allowing it to engage with any cloud-layer \gls{LLM} service that mirrors or bridges to the OpenAI \gls{API}, such as Anthropic's Claude model.

As an additional abstraction, SpeziLLM provides a declarative function-calling \gls{DSL} that simplifies integration, remaining \gls{LLM}-agnostic to ensure consistent application code across different providers (see \autoref{lst:spezillm-function-call}).
It translates function-calling definitions into the appropriate format for specific providers, such as OpenAI, facilitating seamless integration, enhancing code reliability, and bypassing the complexities associated with handling untyped function-calling JSON definitions.

\begin{figure}[h]
    \centering
    \begin{minipage}{\textwidth}
\begin{lstlisting}[
            language=swift,
            linewidth=\linewidth
        ]
struct GetHealthDataFunction: LLMFunction {
    static let name = "get_health_data"
    static let description = "Fetch patient health data based on specified categories"

    @Parameter("Categories of health data")
    var healthDataCategories: [String]

    func execute() async throws -> String? {
        // Use `healthDataCategories` to fetch and return health data as a single string
    }
}
\end{lstlisting}
    \end{minipage}
    \caption{Swift code showcasing the usage of SpeziLLM's declarative function calling \gls{DSL}. The \gls{LLM} function fetches the health data from a patient record based on requested health data categories and returns the data to the \gls{LLM}.}
    \label{lst:spezillm-function-call}
\end{figure}

A full code example is available in the open-source SpeziLLM documentation.\footnote{\url{https://swiftpackageindex.com/stanfordspezi/spezillm/documentation/spezillmopenai/functioncalling}}

\subsection{Fog layer}

SpeziLLM is designed to streamline the complex fog-based, decentralized \gls{LLM} inference system implementation process.
The Swift-based framework (the client) is complemented by a server-side Docker-packaged fog node component, which includes scripts to facilitate the rapid deployment and systemic integration of new fog nodes.
The fog node’s \gls{API} aligns with the OpenAI API, enabling dynamic substitution of the underlying \gls{LLM} inference service.

The client, represented by the SpeziLLM framework, initiates service discovery by advertising fog nodes (\autoref{fig:llm-fog-inference}).
This resource announcement and selection process utilizes \gls{mDNS}~\cite{rfc6762} and \gls{DNS-SD}~\cite{rfc6763}
to ensure widespread compatibility across various operating systems, including Linux through Avahi\footnote{\url{https://avahi.org/}} and Apple platforms via Bonjour\footnote{\url{https://developer.apple.com/bonjour}}.

Upon receiving an inference request, SpeziLLM discovers and resolves available computing resources within the local network to an \gls{IP}v4 or \gls{IP}v6 address.
The client establishes a secure connection to the fog node using \gls{SSL}/\gls{TLS}~\cite{rfc8446} and authenticates using a \gls{JSON} Web Token (JWT)-based token~\cite{rfc7519}.

The \gls{LLM} inference request is performed on the fog node using the open-source Ollama framework\footnote{\url{https://github.com/ollama/ollama}}, which extends the capabilities of the llama.cpp library.
This framework features model persistence management, default configurations, and an \gls{LLM} interface that mirrors the OpenAI \gls{API}.
Ollama facilitates the execution of widely-accessible \glspl{LLM} like Llama3 and Gemma.
A comprehensive list of supported models is available at \footnote{\url{https://ollama.com/library}}.

\subsection{Edge or local layer}
\label{subsec:spezillm-edge}

SpeziLLM's local execution builds on the open-source \textit{llama.cpp} library\footnote{\url{https://github.com/ggerganov/llama.cpp}}, which now supports all major open-weight models.
The library leverages Apple's Silicon hardware acceleration and software frameworks like \textit{Accelerate}\footnote{\url{https://developer.apple.com/documentation/accelerate}} and Metal.
It supports vectorization, quantization, hybrid CPU/GPU inference, and the offloading of specific inference tasks to optimized chip components.
An open-source XCFramework version of llama.cpp~\footnote{\url{https://github.com/StanfordBDHG/llama.cpp}} was forked, enabling binary distribution for all Apple platforms, proper semantic versioning, and dependency management via the \gls{SPM}.

SpeziLLM manages resource tasks for local execution, freeing developers to focus on application logic. The \gls{LLM} \textit{LocalPlatform} (\autoref{sec:architecture}) enforces one sequential execution job at a time, ensuring proper local inference. 
The framework also includes utility components, like an \gls{LLM} download and persistence manager for efficient model file retrieval and setup.

SpeziLLM's local execution layer supports the open-weight language models listed in \autoref{tab:spezillmlocal-supportedmodels} with minimal configuration. Developers must provide a \textit{LocalSchema} configuration and an \gls{LLM} model file in the llama.cpp \gls{GGUF} format\footnote{\url{https://github.com/ggerganov/ggml/blob/master/docs/gguf.md}}.
SpeziLLM's local execution component additionally supports the integration of any \gls{GGUF} format model and configuration of varying prompt structures.

\begin{table}[h]
\centering
\begin{tabular}{@{}lcc@{}}
\toprule
\textbf{Language Model} & \textbf{Variations} & \textbf{Vendor} \\ \midrule
Llama2 & 7B, 13B, 70B (Instruct and Chat) & Meta Platforms \\
Gemma & 2B, 7B & Google \\
Phi-2 & 3B & Microsoft \\
\bottomrule
\end{tabular}
\caption[Natively supported local models by SpeziLLM via the llama.cpp inference environment.]{Natively supported local models by SpeziLLM via the llama.cpp inference environment.}
\label{tab:spezillmlocal-supportedmodels}
\end{table}

The inference capabilities of \glspl{LLM} are rapidly advancing (see \autoref{sec:discussion}).
At present, llama.cpp represents the most efficient mechanism for executing language models on mobile devices.
\subsection{Methods}
\label{sec:methods}

We present six diverse case studies showcasing the application of SpeziLLM in the development of various mobile platforms, each tailored toward distinct digital health objectives (\autoref{tab:spezillm_applications}).
Each selected application used the Llama 2 model (7B variant)~\cite{touvron2023llama} in the local and fog layers.
The local layer used an iPhone 15 Pro with an A17 Pro \gls{SoC} and 8GB RAM, and the fog node used a MacBook Pro 16" with an M1 Pro chip, ten cores, and 16GB RAM.
The fog node ran in a Docker container, which simplified deployment but added performance overhead as a result of limited hardware acceleration.
The cloud layer used OpenAI's GPT-4 (\texttt{gpt-4-0125-preview}).

\begin{table*}[htbp]
\begin{tabular}{@{}|l|p{3.2cm}|p{2cm}|p{1.3cm}|p{1.5cm}@{}|}
\toprule
\textbf{Application} & \textbf{Description} & \textbf{Developers} & \textbf{Models} & \textbf{Status} \\ \midrule
1.
LLMonFHIR & Explains and provides helpful context for \gls{FHIR}-formatted patient data via \glspl{LLM} &
Stanford Biodesign Digital Health & Cloud, Fog, Edge (Llama2 7B) & \Gls{MVP} built, study planned \\
\hline
2. 
OwnYourData & Aims to increase diversity in cancer clinical trials through \gls{LLM} and FHIR EHR-based patient/study matching. & 
Stanford Biodesign Digital Health and OwnYourData LLC & Cloud (OpenAI GPT-4) & In development \\
\hline
3. 
HealthGPT & Enables users to query and interact with their health data stored in Apple Health using natural language. &
Stanford Biodesign Digital Health & Cloud, Fog, Edge (Llama2 7B) & \Gls{MVP} built \\
\hline
4. 
Nourish & Meal tracking app designed for outpatient support of individuals with Avoidant/Restrictive Food Intake Disorder (ARFID). & 
Stanford Biodesign Digital Health \& Lucile Packard Children's Hospital Stanford & Cloud (OpenAI GPT-4) & \gls{MVP} built; study planned \\
\hline
5. 
Stronger & Tracks protein intake and resistance exercise training in postmenopausal research participants. &
Stanford Biodesign Digital Health \& Stanford Medicine & Cloud (OpenAI GPT-4) & \Gls{MVP} built, study planned \\
\hline
6. 
Intake & Pre-populates medical intake forms based on FHIR records via interactive \glspl{LLM}. & 
Stanford Biodesign Digital Health & Cloud (OpenAI GPT-4) & \Gls{MVP} built, study planned \\
\bottomrule
\end{tabular}
\caption[A list of six health applications built between 2023 and 2024 using SpeziLLM.]{A list of six health applications built between 2023 and 2024 using SpeziLLM.}
\label{tab:spezillm_applications}
\end{table*}

\begin{table}[ht]
    \centering
    \renewcommand{\arraystretch}{1.2}
    \begin{tabular}{|m{0.45cm}|m{10.6cm}|}
        \hline
        \textbf{ID} & \textbf{Question} \\
        \hline
        Q1 & How easy was it to integrate SpeziLLM into your iOS application? (1 = Very hard, 5 = Very Easy)\\
        \hline
        Q2 & Rate the learning curve of SpeziLLM. (1 = Very Steep, 5 = Very Gentle)\\
        \hline
        Q3 & Rate your overall satisfaction with SpeziLLM. (1 = Not satisfied, 5 = Very satisfied)\\
        \hline
        Q4 & How intuitive did you ﬁnd the API naming conventions and architecture? (1 = Very Unintuitive, 5 = Very Intuitive)\\
        \hline
        Q5 & How effective was SpeziLLM in simplifying the use of LLMs in your application? How significant was the improvement in development speed? (1 = Very Ineffective, 5 = Very Effective)\\
        \hline
        Q6 & To what extent did the provided documentation help you in implementing and use the APIs and DSLs (Domain-Speciﬁc Language) of SpeziLLM? (1 = Not at all, 5 = Very much)\\
        \hline
        Q7 & How well did SpeziLLM handle errors and exceptions? How comprehensive are the error messages and debugging information provided by the framework? (1 = Not at all, 5 = Very well)\\
        \hline
        Q8 &
        To what extent does SpeziLLM offer innovative features not available in other similar tools? How would you rate the SpeziLLM in comparison to other libraries and frameworks (also outside the Swift ecosystem) you used for LLM-based interactions? (1 = Not at all, 5 = Lots of)\\
        \hline
    \end{tabular}
    \caption[Survey questions for CS342 students.]{Survey questions for CS342 students who used SpeziLLM to develop their iOS applications about SpeziLLM's usability, intuitiveness, and adaptability.}
    \label{table:cs342-questions}
\end{table}

We also evaluated SpeziLLM's utility in mobile application development for those without prior experience with SpeziLLM or the Stanford Spezi ecosystem. To that end, CS students enrolled in CS342\footnote{\url{https://cs342.stanford.edu}} at Stanford University were voluntarily sampled.
CS342 is a ten-week, team-based course in which students learn to design and build secure digital solutions to unmet health needs.
An anonymous survey was distributed via Google Forms to 15 students whose final applications (\textit{Nourish} (4), \textit{Stronger} (5), \textit{Intake} (6)) were developed using SpeziLLM in groups of five.
The survey consisted of eight five-point Likert-scale questions (detailed in \autoref{table:cs342-questions}), and was designed to assess the usability and functionality of SpeziLLM.
Data collection spanned five days. Participants were notified via direct messages (with one "reminder") to maximize response rates. Responses were analyzed using descriptive statistics.

The primary objective of this study was to assess the usability, functionality, and adaptability of SpeziLLM among novices in Swift and mobile app development.
Limitations include a small sample size, poor generalizability (the sample is composed of Computer Science students in a particular course at a particular institution), the lack of a control group, and varying levels of respondent engagement with SpeziLLM, the broader Stanford Spezi ecosystem, and the Swift language.
\subsection{Results}
\label{sec:results}

The case studies shown in \autoref{tab:spezillm_applications} showcase SpeziLLM's versatility across various digital health use cases.
\textit{LLMonFHIR} (1), an open-source iOS application~\cite{schmiedmayer2023llmonfhir}
that facilitates interactive dialogue between users and their \gls{FHIR} records~\cite{schmiedmayer2024llmonfhir},
benefited greatly from the integration of SpeziLLM's \gls{LLM} capabilities and prebuilt UI elements, particularly the \gls{API} token capture and chat components.
Application of the SpeziLLM framework enabled the adoption of OpenAI's function calling declarative \gls{DSL} (see \autoref{fig:spezillm-function-calling}), reducing code complexity and improving performance via parallel processing of function calls.

\begin{figure}[h]
    \centering
    \begin{subfigure}[t]{0.32\linewidth}
        \centering
        \includegraphics[width=\textwidth]{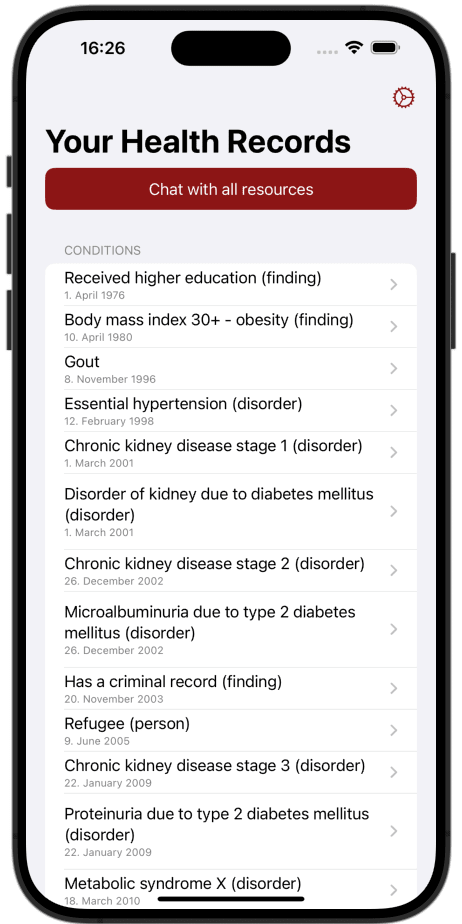}
        \caption{Overview of all available FHIR records.}
        \label{fig:llm_fhir_overview}
    \end{subfigure}
    \hfill
    \begin{subfigure}[t]{0.32\linewidth}
        \centering
        \includegraphics[width=\textwidth]{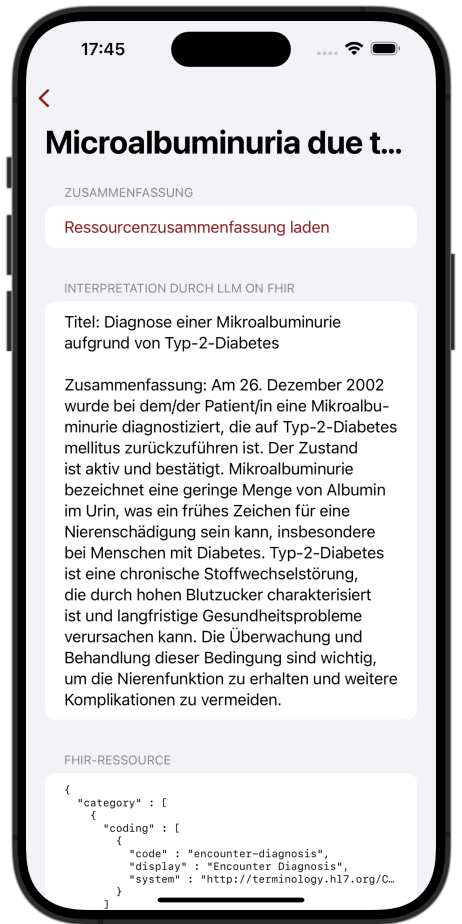}
        \caption{Interpretation and Summary of a single resource.}
        \label{fig:llm_fhir_interpretation}
    \end{subfigure}
    \hfill
    \begin{subfigure}[t]{0.32\linewidth}
        \centering
        \includegraphics[width=\textwidth]{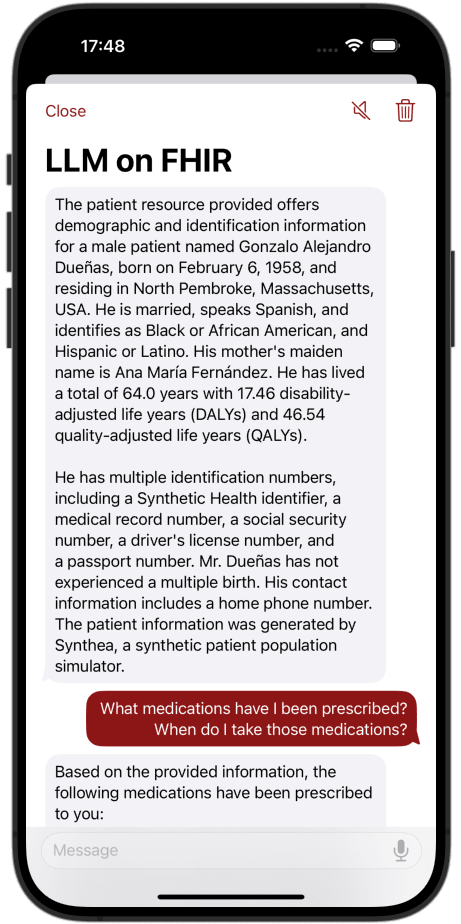}
        \caption{Interactive Chat with all FHIR health records.}
        \label{fig:llm_fhir_chat}
    \end{subfigure}
    \caption{Screenshots of the LLMonFHIR iOS Application.}
    \label{fig:llm_fhir_screenshots}
\end{figure}

This implementation exhibits SpeziLLM's uniform interface for \gls{LLM} interactions, which uses function calling to retrieve \gls{FHIR} resources and enable resource summarization--a functionality that has been extracted into the \textit{Spezi FHIR}~\cite{schmiedmayer2023spezifhir}
Swift package (also used by \textit{OwnYourData} (2)).

LLMonFHIR (\autoref{fig:llm_fhir_screenshots}) can switch between underlying \gls{LLM} inference environments, combining different \gls{LLM} layers to dynamically dispatch tasks to the most suitable \gls{LLM}.
Simple tasks like the summarization, transformation, or interpretation of single, raw \gls{FHIR} resources (\autoref{fig:llm_fhir_interpretation}) can be handled by local or fog-based models with higher trust levels (\autoref{sec:architecture}), minimizing exposure of potential patient identifiers to remote cloud providers.
Outputs from these tasks serve as inputs for more complex tasks, such as the interactive chat view (\autoref{fig:llm_fhir_chat}), which builds on locally-generated summaries, allowing the cloud-based OpenAI \gls{LLM} to fetch, explain, and interpret the summaries in a user dialogue without accessing raw, potentially sensitive \gls{FHIR} resources.

\begin{figure}[h]
    \centering
    \begin{subfigure}[t]{0.32\linewidth}
        \centering
        \includegraphics[width=\textwidth]{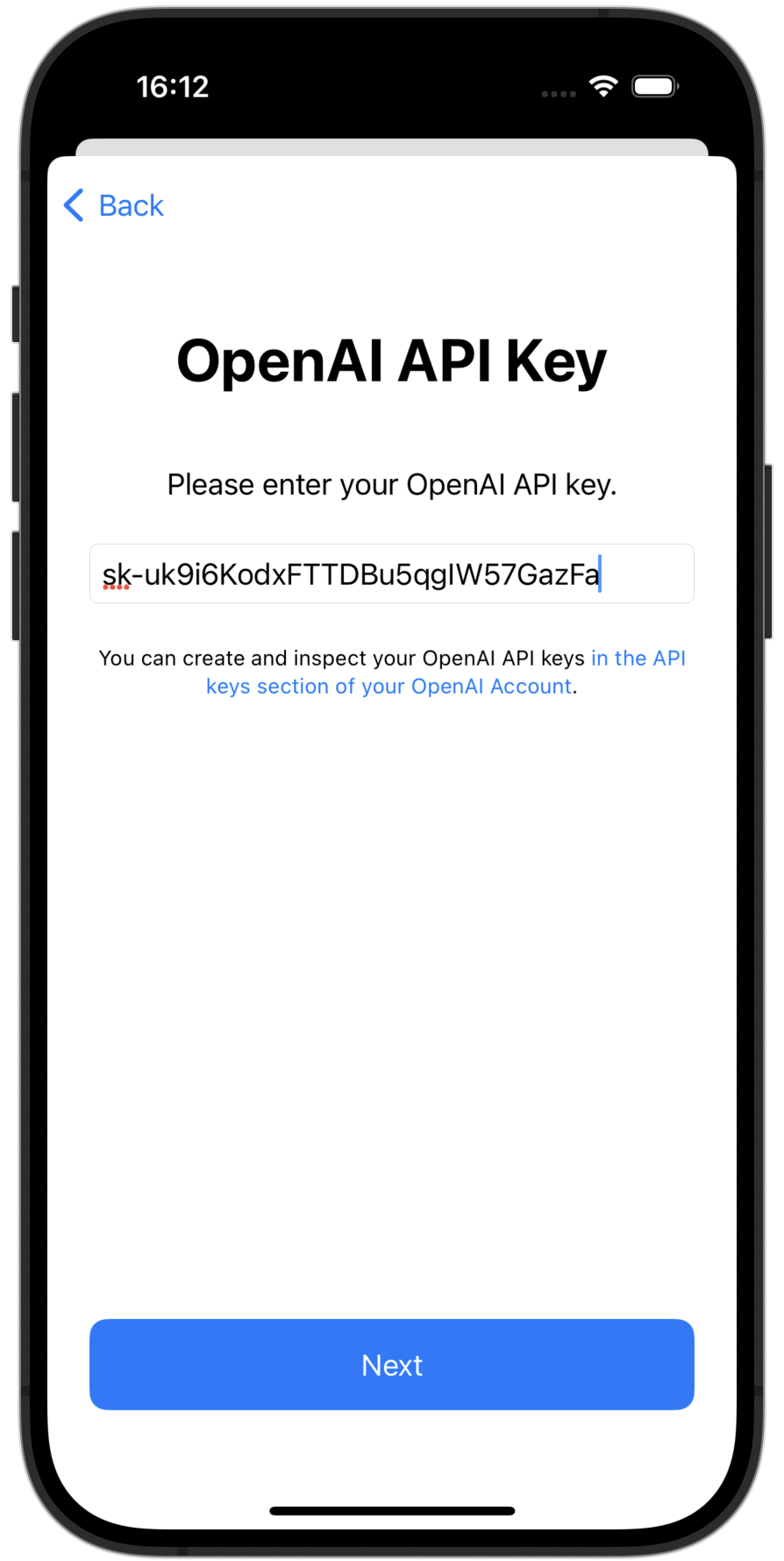}
        \caption{Collect OpenAI API Key during Onboarding.}
        \label{fig:healthgpt_key}
    \end{subfigure}
    \hfill
    \begin{subfigure}[t]{0.32\linewidth}
        \centering
        \includegraphics[width=\textwidth]{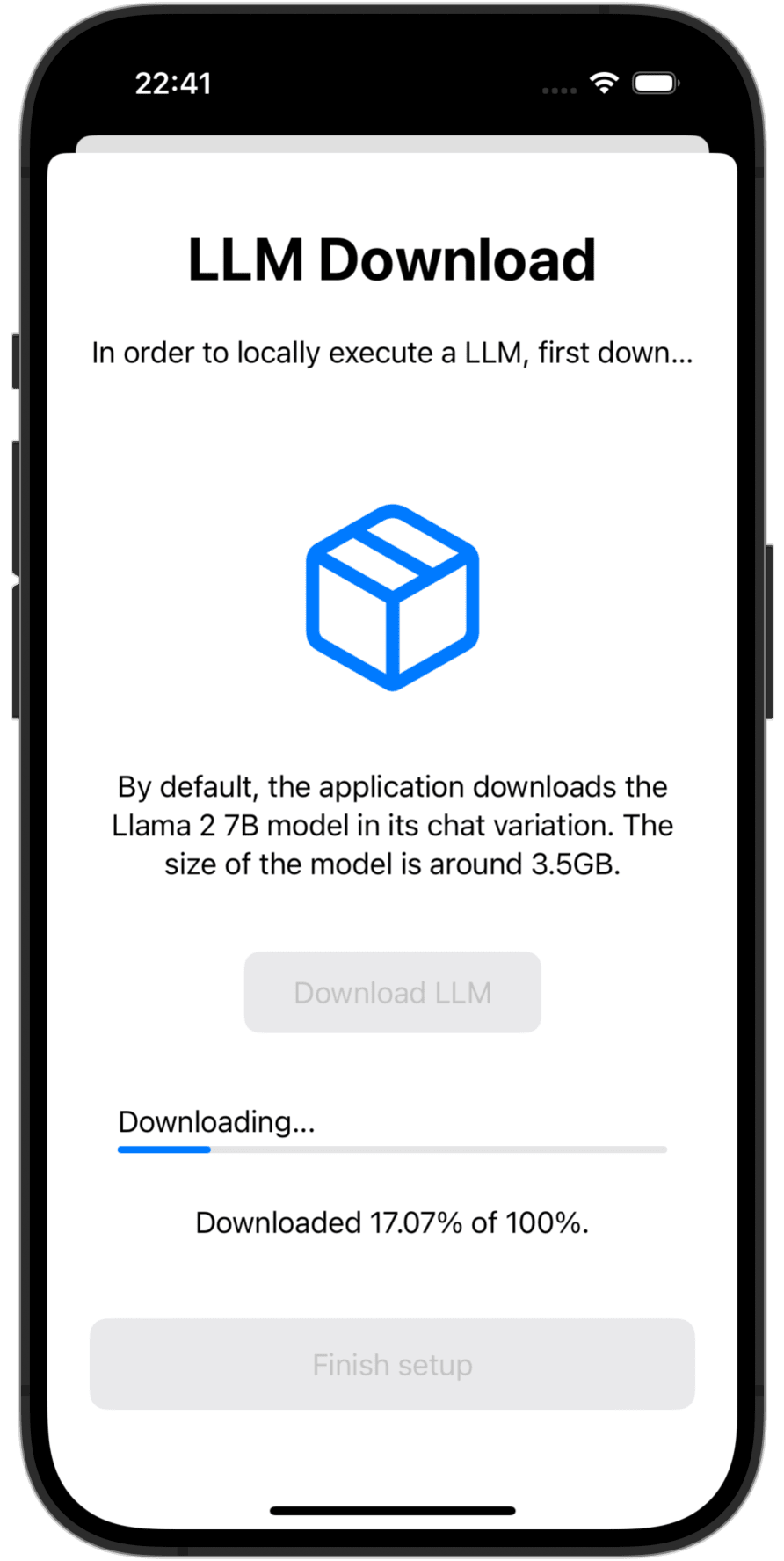}
        \caption{Download the to-be-used local \gls{LLM}.}
        \label{fig:healthgpt_model}
    \end{subfigure}
    \hfill
    \begin{subfigure}[t]{0.32\linewidth}
        \centering
        \includegraphics[width=\textwidth]{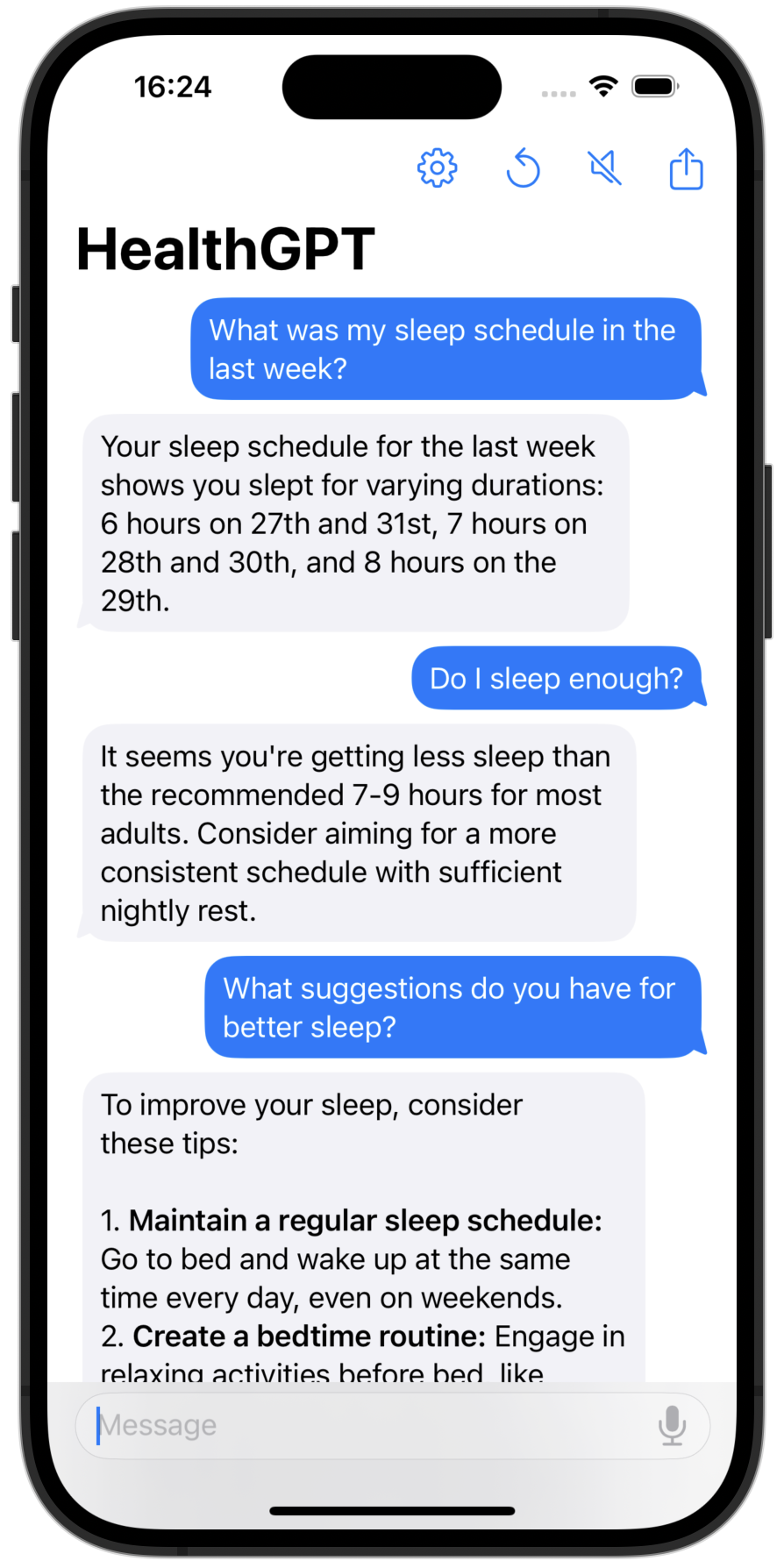}
        \caption{Chat with Apple Health records.}
        \label{fig:healthgpt_chat}
    \end{subfigure}
    \caption[Screenshots of the HealthGPT iOS Application.]{Screenshots of the HealthGPT iOS Application.}
    \label{fig:healthgpt_screenshots}
\end{figure}

SpeziLLM's uniform interface and cross-layer dynamic dispatch capability are heavily utilized in the \textit{HealthGPT} (3) application (\autoref{fig:healthgpt_screenshots}), rearchitected to utilize SpeziLLM and multiple \gls{LLM} execution environments~\cite{schmiedmayer2023healthgpt}.
This open-source, Stanford Spezi-based application allows users to interactively query their Apple Health data using natural language~\cite{schmiedmayer2023healthgpt}.
Users can query metrics like sleep, step count, exercise minutes, body mass, and heart rate through a bidirectional speech-to-text capable chat interface at any degree of complexity and in various languages (\autoref{fig:healthgpt_chat}).
Aside from the OpenAI cloud LLM inference (\autoref{fig:healthgpt_key}), HealthGPT supports models running in the local or fog layer (\autoref{fig:healthgpt_model}), ensuring that sensitive health information and user conversations are processed in trusted environments.

\begin{figure}[h]
\centering
\includegraphics[width=0.8\linewidth]{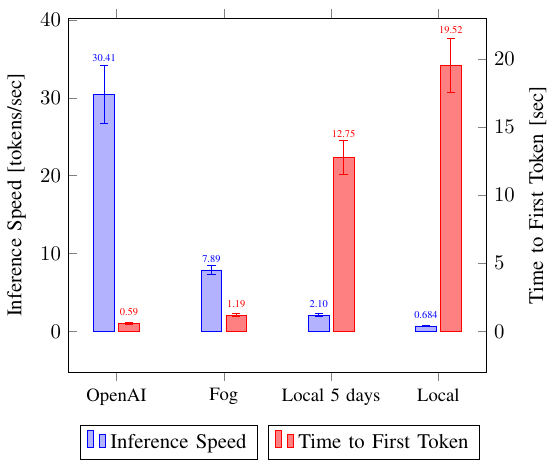}
\caption[Comparison of inference speed and time to first token for different LLMs in HealthGPT.]{Comparison of inference speed and time to first token for different LLMs in HealthGPT.
Each measurement represents the mean of five quantifications, with the standard deviation between these measurements in brackets.
Measurements were made in response to the question "\textit{How much did I sleep last week and how can I improve my sleep?}".}
\label{fig:healthgpt-model-performance-chart}
\end{figure}

In developing these applications, we found that the local and fog layer output was comprehensible, but often more verbose and less detailed than OpenAI's GPT.
We observed a notable delay in response times from the local layer model, which we attributed to limited computational resources (e.g., modern mobile devices' restricted main memory) (\autoref{sec:background}).
To improve response time, we capitalized on an observed inverse correlation between context size and local inference speeds, reducing the context size in the local \gls{LLM} from 14 days to five.
The fog layer \gls{LLM}, despite being containerized with Docker, achieved near-natural output generation speeds akin to 200-300 words per minute (five to seven tokens per second)\footnote{https://scholarwithin.com/average-reading-speed}.

SpeziLLM has been additionally instantiated in the Stanford
University CS342 projects \textit{Nourish} (4), \textit{Stronger} (5)~\cite{schmiedmayer2024stronger}, and \textit{Intake} (6)~\cite{schmiedmayer2024intake}, all of which used the framework for chat interfaces and function calling mechanisms to interface with application domain logic.
The Intake project populates medical intake forms with \gls{EHR}-provided \gls{FHIR} record data via interactive language models.
The selection and filtering of relevant \gls{FHIR} records combines rules-based logic with \gls{LLM} inference, addressing the limitations of rules-based logic alone in handling the complexities and customizations of the \gls{FHIR} format. Additional user-friendly features include the collection of additional patient data (the primary concern/reason for visit, for example) through an interactive chat-based \gls{LLM} interface.

\begin{figure}[h]
\centering
\includegraphics[width=0.8\linewidth]{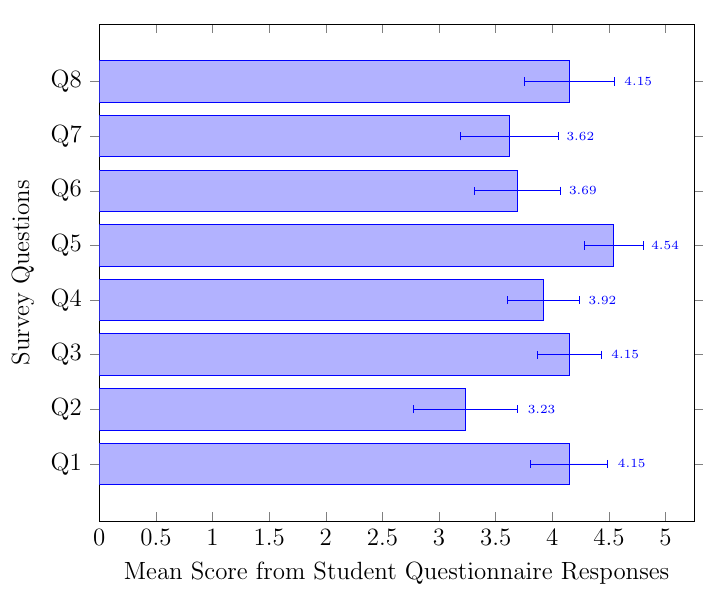}
\caption[Evaluation of SpeziLLM's integration, usability, and innovation factors by CS342 students.]{Evaluation of SpeziLLM's integration, usability, and innovation factors by CS342 students. The mean and standard deviation derived from the responses of 13 participants are shown.}
\label{fig:cs342_evaluation-chart}
\end{figure}

The student survey portion of our analysis had an 86.67\% response rate (n = 13) to all eight questions.
Quantitative reviews of SpeziLLM were generally positive (\autoref{fig:cs342_evaluation-chart}). Students noted the simplicity of integrating the framework into their digital health prototypes (Q1) and praised its utility in easing the use of \glspl{LLM} therein (Q5).
The functionality, including the transparent execution environment and declarative OpenAI function calling \gls{DSL}, received similarly high ratings (Q8).
Respondents reported--albeit with the highest variation--that the learning curve was challenging (Q2).
SpeziLLM's API design, naming conventions, and documentation were thought to be intuitive and helpful (Q4, Q6).
The framework's error and exception handling capability was identified as needing improvement (Q7).
Overall, satisfaction was high, with students noting that SpeziLLM significantly eased and accelerated \gls{LLM} integration into their digital health applications (Q3).
The time-constrained eight-week course coding period necessitated a steep learning curve, and negative feedback stemmed largely from limited software engineering experience and a lack of prior exposure to Swift, SwiftUI, and the Stanford Spezi ecosystem.

\section{Discussion}
\label{sec:discussion}

\Glspl{LLM} have the capacity to transform vast, unstructured data from \glspl{EHR}, wearable devices, and digital health applications into actionable insights, improving individual health and health care delivery systems alike ~\cite{schmiedmayer2024llmonfhir, clusmann2023future, thirunavukarasu2023large, shah2023creation}.
Where their use involves the transmission of patient data to remote clouds, however, valid privacy, trust, and cost concerns arise (\autoref{sec:introduction}).
To that end, SpeziLLM is designed to move \gls{LLM} computations closer to the user using a decentralized, layered fog computing model.
SpeziLLM's uniformity facilitates the division of extensive \gls{LLM} tasks into smaller segments, which are assessed for complexity and assigned to execution layers accordingly.
The output generated by a more trusted layer serves as input for more capable, less trusted layers, pre-processing sensitive health data prior to its exposure to opaque cloud \gls{LLM} providers.

Throughout the development of our case studies, we observed an inverse correlation between context size and inference speed. The memory swapping associated with larger contexts served to degrade performance, even using advanced devices like the iPhone 15 Pro with 8GB RAM~\cite{xu2023llmcad} (\autoref{sec:results}).
Similar reductions occur in fog nodes within Docker containers.
The high cost of inference-capable hardware may pose a challenge to the widespread adoption of \gls{LLM} capabilities.

We found that smaller, locally-executed \glspl{LLM} handle straightforward tasks (e.g., data transformation and summarization) well, but are less effective at generating new insights than larger models, often producing verbose outputs and exhibiting weaker contextual understanding.
Methods like few-shot prompting~\cite{brown2020language} can enhance performance, underscoring the value of an approach that leverages multiple different models and computational layers in accordance with their individual strengths and limitations.

Technological advances continue to create the potential for local \gls{LLM} execution on edge devices.
Development of \gls{LLM} inference software optimizations for mobile operating systems is ongoing, with a particular focus on improving execution in memory-limited contexts.
These projects, which include Android's AICore and Apple's CoreML / Intelligence, aim to optimize large-scale model execution on standard mobile devices and with extensive contexts~\cite{gunter2024appleintelligence, alizadeh2024llm, mckinzie2024mm1, moniz2024realm}.
SpeziLLM's flexible model for uniform interactions can readily incorporate advancements in local execution capability, accelerating \gls{LLM} application development and deployment for digital health innovators and the patients they serve.
\section{Conclusion}
\label{sec:conclusion}

This work demonstrates the potential of our open-source, dynamic, and uniform inference framework, SpeziLLM, to enhance the execution of LLM tasks across edge, fog, and cloud environments. 
By shifting LLM execution closer to the user’s device within a decentralized fog computing architecture, SpeziLLM addresses critical privacy, trust, and financial concerns related to cloud-based solutions, all while maintaining robust LLM functionalities near the data source.

SpeziLLM’s architecture encompasses three tiers: local/edge, fog, and cloud. 
The framework enables efficient local inference on Apple's mobile platforms, establishes a decentralized fog layer with discoverable LLM nodes, and seamlessly integrates with the OpenAI API through a declarative DSL for function calling. 
SpeziLLM’s instantiation across six mobile applications highlights its versatility and utility in an array of digital health contexts.

SpeziLLM's primary contribution is its uniform, LLM-agnostic interface, which facilitates transparent transitions between inference environments. 
This approach offers a viable solution to the challenges of LLM execution in sensitive and resource-constrained environments \cite{gunter2024appleintelligence, alizadeh2024llm, mckinzie2024mm1, moniz2024realm}, facilitating the segmentation of extensive LLM tasks into smaller components and the distribution of such segments across complexity- and trust-matched layers. 
Lightweight models, such as Llama 2, can efficiently handle simpler tasks within the local and fog layers, while more complex tasks can be processed in the cloud. Future developments in efficient \gls{LLM} execution with limited memory will further refine SpeziLLM's performance and applicability.

\section*{Achknowledgements}

The research was partially funded by a German Academic Exchange Service (DAAD) "Internationale Forschungsaufenthalte für Informatikerinnen \& Informatiker" scholarship (91886835) and the Bavaria California Technology Center (BaCaTec) project fund (Nr. 10 2023-2).
We thank the Stanford Mussalem Center for Biodesign for supporting this project and our digital health research.

\bibliographystyle{IEEEtran}
\bibliography{IEEEabrv,bibliography}

\end{document}